\documentclass{article}
\PassOptionsToPackage{numbers, compress}{natbib}

\usepackage[preprint]{neurips_2021}

\usepackage[utf8]{inputenc} 
\usepackage[T1]{fontenc}    
\usepackage{hyperref}       
\usepackage{url}            
\usepackage{booktabs}       
\usepackage{amsfonts}       
\usepackage{nicefrac}       
\usepackage{microtype}      
\usepackage{xcolor}         
\usepackage{tikz}
\usetikzlibrary{shapes.geometric}
\usepackage{pgfplots}
\pgfplotsset{compat=newest}
\pgfdeclareplotmark{mystar}{
\node[star,star point ratio=2.25,minimum size=6pt,
      inner sep=0pt,draw=black,solid,fill=red] {};
}

\usepackage{subcaption}
\usepackage{multirow}
\usepackage{algorithm}
\usepackage{algorithmicx}
\usepackage{algpseudocode}
\usepackage{booktabs}
\usepackage{amssymb}
\usepackage{amsmath}
\usepackage{todonotes}

\usepackage{siunitx}
\usepackage{wrapfig}
\usepackage{lipsum}
\newcommand*\samethanks[1][\value{footnote}]{\footnotemark[#1]}

\title{Early Exiting with Ensemble Internal Classifiers}

\author{%
  Tianxiang Sun$^1$\thanks{Equal contribution.}~~~~~~~~
  Yunhua Zhou$^1$\samethanks~~~~~~~~
  Xiangyang Liu$^1$~~~~~~~
  \\
  \bf 
  Xinyu Zhang$^2$~~~~~~
  Hao Jiang$^2$~~~~~~
  Zhao Cao$^2$~~~~~~
  Xuanjing Huang$^1$~~~~~~
  Xipeng Qiu$^1$\thanks{Corresponding author.}\\
  $^1$School of Computer Science, Fudan University~~~
  $^2$Huawei Poisson Lab\\
  \tt \{txsun19,zhouyh20,xiangyangliu20,xjhuang,xpqiu\}@fudan.edu.cn\\
  \tt \{zhangxinyu35,jianghao66,caozhao1\}@huawei.com \\
}

\begin{document}

\maketitle

\begin{abstract}
  As a simple technique to accelerate inference of large-scale pre-trained models, early exiting has gained much attention in the NLP community. It allows samples to exit early at internal classifiers without passing through the entire model. Most existing work usually trains the internal classifiers independently and employs an exiting strategy to decide whether or not to exit based on the confidence of the current internal classifier. However, none of these works takes full advantage of the fact that the internal classifiers are trained to solve the same task therefore can be used to construct an ensemble. In this paper, we show that a novel objective function for the training of the ensemble internal classifiers can be naturally induced from the perspective of ensemble learning and information theory. The proposed training objective consists of two terms: one for accuracy and the other for the diversity of the internal classifiers. In contrast, the objective used in prior work is exactly the accuracy term of our training objective therefore only optimizes the accuracy but not diversity. Further, we propose a simple voting-based strategy that considers predictions of all the past internal classifiers to infer the correct label and decide whether to exit. Experimental results on various NLP tasks show that our proposed objective function and voting-based strategy can achieve better accuracy-speed trade-offs.
\end{abstract}

\section{Introduction}

Recent years have witnessed great success of large-scale pre-trained language models (PLMs)~\cite{Devlin2019BERT,Raffel2020T5,Lan2020ALBERT,Qiu2020survey}. Despite their significant improvement in accuracy, the inference is time-consuming and computationally expensive.
To tackle this problem, \textit{early exiting} has emerged as a simple yet effective method to achieve efficient inference~\cite{Xin2020DeeBERT,Liu2020FastBERT,Schwartz2020Right,Zhou2020BERT}.




Early exiting methods usually add internal classifiers to different layers of a model. By training these internal classifiers with the ground truth, the model has a chance to predict the correct label and exit earlier during inference. To be summarized, early exiting methods have two steps: \textbf{(a)} Training the internal classifiers on downstream tasks to make them capable of making predictions, \textbf{(b)} Designing an exiting strategy to decide whether to exit early or continue to the next layer.

\begin{figure}
    \centering
    \begin{subfigure}{0.45\linewidth}
    \centering
    \includegraphics[scale=0.45]{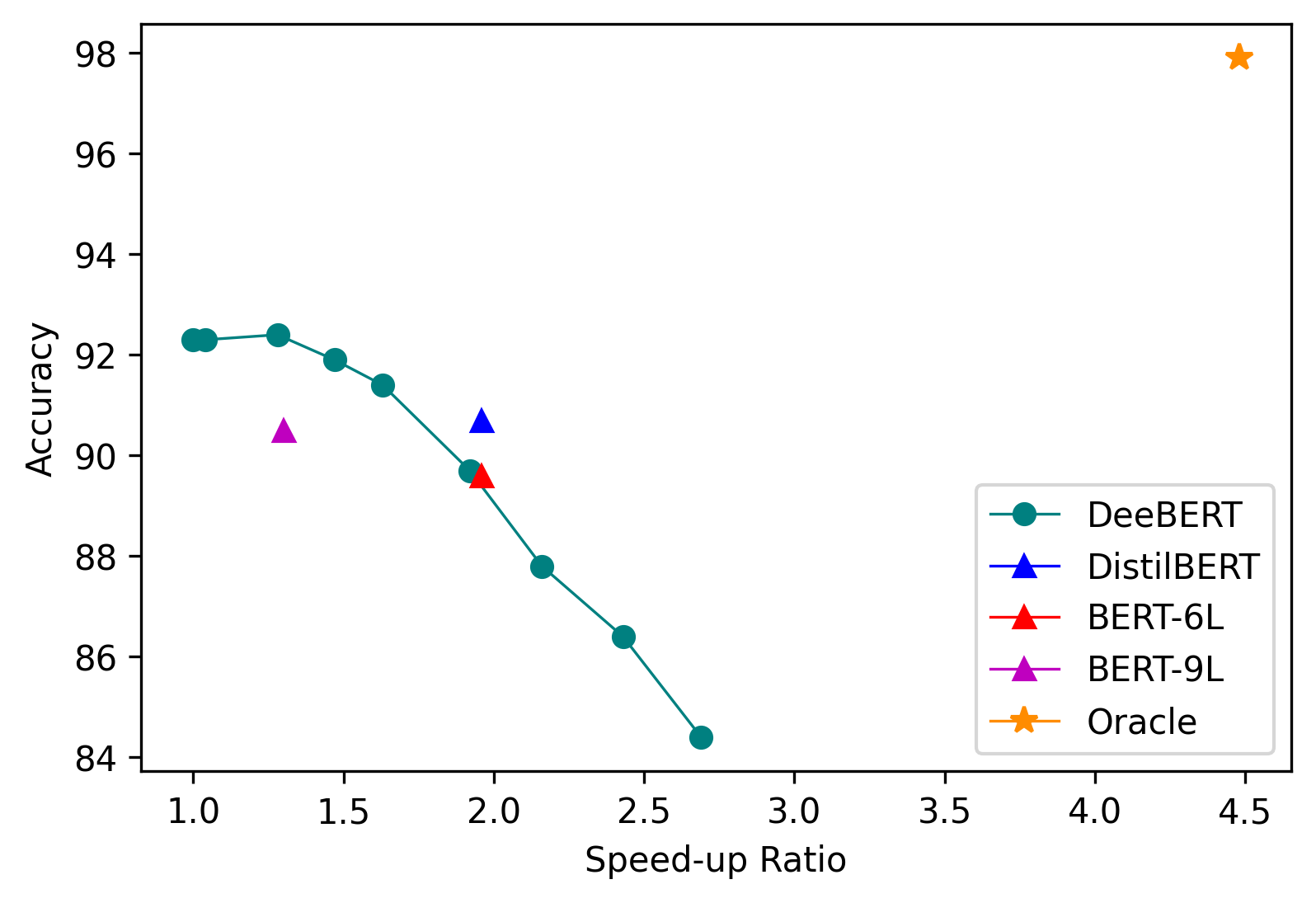}
    \caption{BERT.}
    \end{subfigure}\hfill
    \begin{subfigure}{0.45\linewidth}
    \centering
    \includegraphics[scale=0.45]{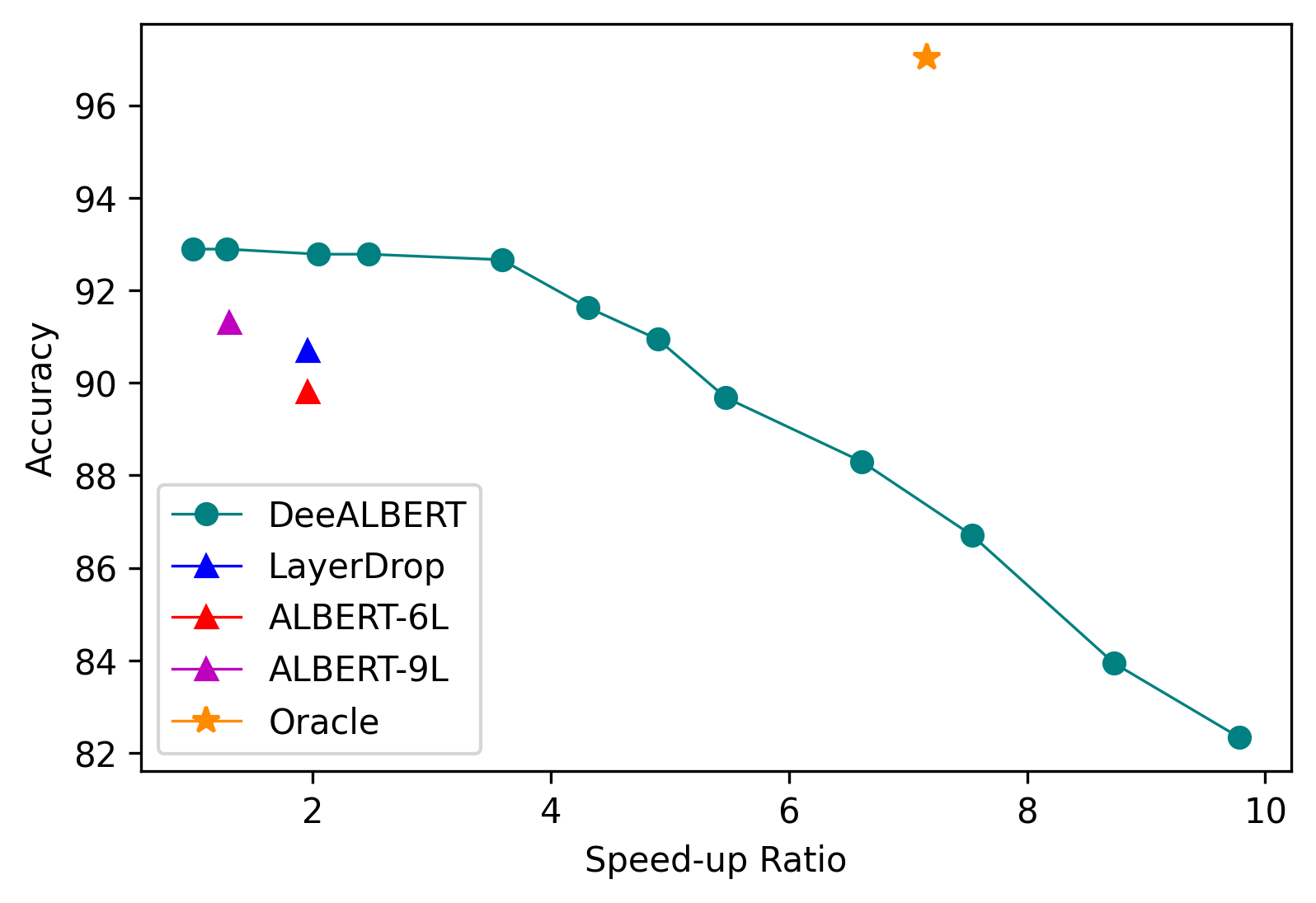}
    \caption{ALBERT.}
    \end{subfigure}
    \caption{Accuracy-speed trade-offs of BERT and ALBERT on SST-2 development set. The results of the static methods (DistilBERT, BERT-6/9L, LayerDrop, ALBERT-6/9L) are taken from \cite{Zhou2020BERT}.}
    \vskip -0.1in
    \label{fig:oracle}
\end{figure}

Most existing work train the internal classifiers independently, and adopt an exiting strategy based on the confidence (e.g., entropy, max probability) of the current single internal classifier. They can significantly accelerate model inference with minimal loss in accuracy. However, we argue that the current methods are still far from the ceiling. To demonstrate the ceiling of early exiting, we define the \textit{Oracle} as a model that always exits at the first internal classifier that correctly predicts the label. Oracle is the smallest (also the fastest) model that correctly solves a given sample. As shown in Figure~\ref{fig:oracle}, compared with the full model, oracle is not only faster but also more accurate since it mitigates the overthinking problem~\cite{Kaya2019Shallow}. Such a tempting oracle motivates us to ask: How can we approach such a fast and accurate model?



As a response, this work attempts to fully exploit the information inside all the available internal classifiers to predict the label and decide whether to exit. To that end, we formulate the early exiting from the perspective of ensemble learning because:
(1) The internal classifiers are trained to solve the same task, which makes them possible to be used to construct an ensemble.
(2) Ensemble methods have shown great power to boost performance by combining predictions of multiple classifiers~\cite{Hansen1990NNE,Schapire1990Strength,Opitz1999Popular}. 
In general, we employ all of the passed internal classifiers to construct an ensemble in a sequential style, which is then used to make predictions and decide whether or not to exit. From the perspective of ensemble learning, to construct a good ensemble, the internal classifiers should be as more accurate as possible, and also as more \textit{diverse} as possible. With the help of information theory, we show that a novel objective function for the training of the ensemble internal classifiers can be naturally induced. The training objective consists of two terms, one is for accuracy and the other is for diversity. Besides, we find that the accuracy term in our training objective is exactly the loss function used in prior work~\cite{Elbayad2020Depth,Schwartz2020Right,Xin2020DeeBERT,Zhou2020BERT}. In contrast, our proposed training objective not only maximizes likelihood of ground truth but also encourages diversity among the internal classifiers. Our formulation shows that the objective to be maximized is a lower bound on the mutual information between the joint probability of the internal classifiers and the ground truth.

The proposed training objective is compatible with any ensemble-based exiting strategies such as the patience-based strategy~\cite{Zhou2020BERT}, which exploits the agreement of the last $s$ internal classifiers as the criterion to make decision. We also propose a simple voting-based strategy that lets all of the past internal classifiers take a vote to infer the answer and decide whether to exit. 
Experimental results on various NLP tasks confirm the effectiveness of our objective function and the voting-based strategy.



\section{Related Work}
Our work is closely related to the research of improving the efficiency of deep neural networks. We categorize the existing methods into two streams: model compression and conditional computation.

\paragraph{Model Compression.} Compressing a cumbersome model to reduce the number or precision of parameters is a straightforward and effective solution. Currently, there are several approaches to achieve model compression: (1) \textit{model pruning}, which removes part of neural network (e.g., weights, neurons, layers, attention heads) that is less important~\cite{Gordon2020Compressing}, (2) \textit{knowledge distillation}, which learns a compact student model that learns from the prediction distributions from the cumbersome teacher model~\cite{Sanh2019DistilBERT,Jiao2020TinyBERT}, (3) \textit{weight sharing} across different parts (e.g., layers) of the model~\cite{Lan2020ALBERT}, (4) \textit{quantization}, which uses low bit precision for parameter storage and speed-up inference with low bit hardware operations~\cite{Shen2020QBERT}, and (5) \textit{module replacing}, which replaces the modules of large-scale models with more compact substitutes~\cite{Xu2020BERT}.


\paragraph{Conditional Computation.} Instead of pursuing a more compact static model, conditional computation is to selectively activate only parts of the model conditioned on a given input~\cite{Bengio2013Estimating,Davis2013Lowrank}. \cite{Graves2016Adaptive} developed an end-to-end halting mechanism, Adaptive Computation Time (ACT), to perform input-adaptive computation, which is later used in Universal Transformer~\cite{Dehghani2019Universal}.

\begin{wraptable}{r}{.5\textwidth}
\centering
\resizebox{.5\textwidth}{!}{
\begin{tabular}{l|rr}
\toprule
\multirow{2}{*}{\textbf{Method}} & \textbf{Training}  & \textbf{Exiting} \\
                                 & \textbf{Objective} & \textbf{Strategy} \\
\midrule
DeeBERT~\cite{Xin2020DeeBERT}                 & $\mathcal{L}_{\text{rel}}$  & $\text{Entropy}(X_l)$                  \\
RightTool~\cite{Schwartz2020Right}               & $\mathcal{L}_{\text{rel}}$  & $\text{MaxProb}(X_l)$               \\
FastBERT~\cite{Liu2020FastBERT}                & $\mathcal{L}_{\text{self-distill}}$  & $\text{Entropy}(X_l)$            \\
PABEE~\cite{Zhou2020BERT}                   & $\mathcal{L}_{\text{rel}}$  & $\text{Agree}(X_{l-s+1:l})$ \\
\midrule
\textit{Ours}           & $\mathcal{L}_{\text{rel}} + \mathcal{L}_{\text{div}}$  & $\text{Voting}(X_{1:l})$         \\
\bottomrule
\end{tabular}}
\caption{Comparison of several early exit methods.}
\label{tab:related}
\end{wraptable}

Early exiting can also be regarded as an instance of conditional computation. It is first applied in computer vision, such as BranchyNet~\cite{Teerapittayanon2016BranchyNet} and Shallow-Deep Network~\cite{Kaya2019Shallow}. Very recently, as the emergence of large-scale models for natural language processing, early exiting is also used to speed up inference of transformer-based models, such as Depth-Adaptive Transformer~\cite{Elbayad2020Depth}, DeeBERT~\cite{Xin2020DeeBERT}, FastBERT~\cite{Liu2020FastBERT}, RightTool~\cite{Schwartz2020Right}, and PABEE~\cite{Zhou2020BERT}. In Table~\ref{tab:related}, we compare our method with these previous work on early exiting mainly from the aspects of training objective and exiting strategy.

\section{An Ensemble Perspective on Early Exiting}

\begin{figure}[ht]
    \centering
    \includegraphics[width=.9\linewidth]{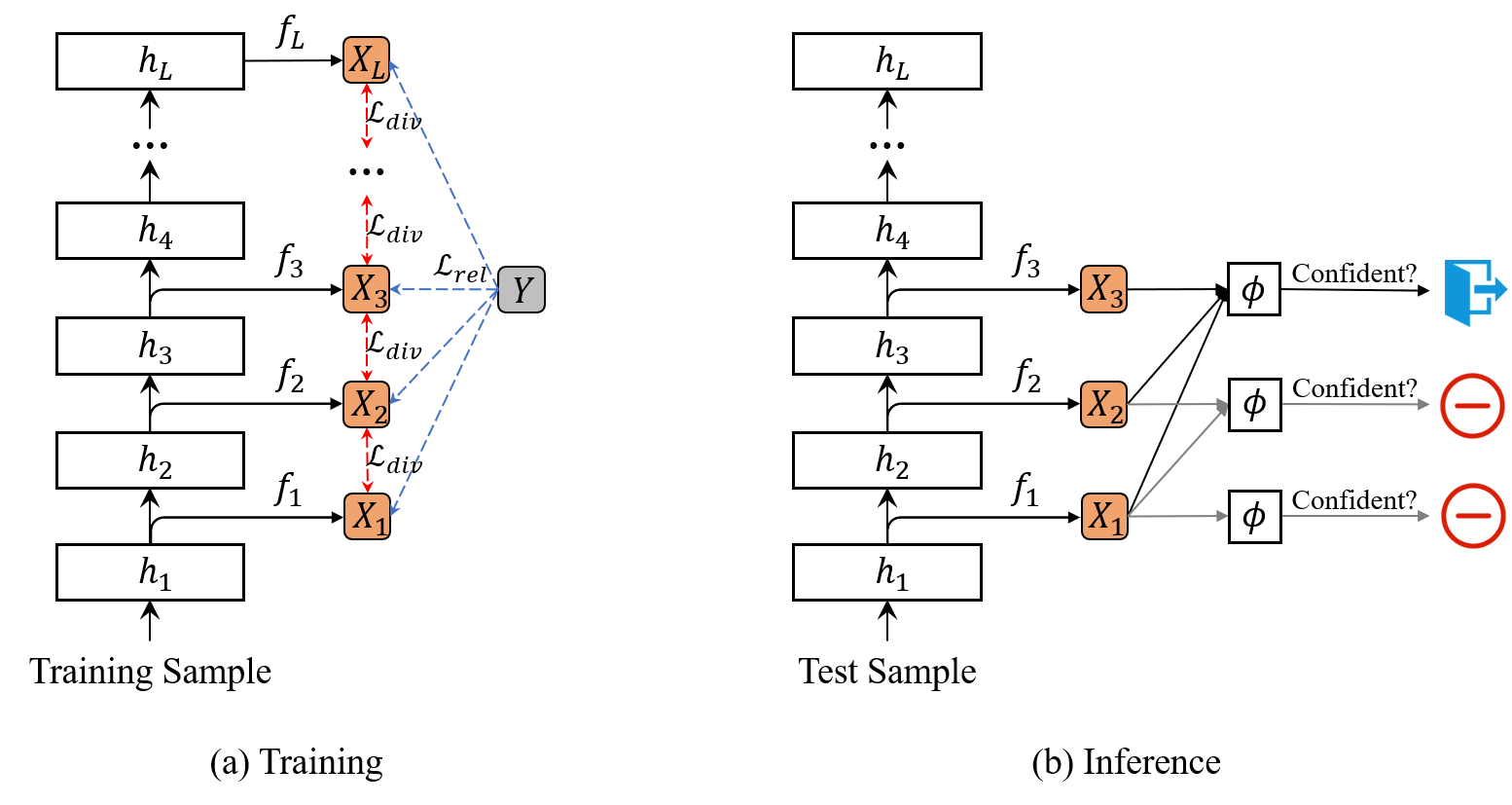}
    \caption{A general illustration of our method. (a) During training, the loss function consists of the relevancy loss $\mathcal{L}_{\text{rel}}$ that maximizes the likelihood of the ground truth $Y$, and the diversity loss $\mathcal{L}_{\text{div}}$ that encourages the diversity of the internal classifiers. (b) During inference, the predictions of each internal classifiers are sequentially collected to be combined by a voting function $\phi$ to decide whether or not to exit early.}
    \label{fig:overview}
\end{figure}

Given a pre-trained language model (PLM) with $L$ layers, denote the hidden state at layer $i$ as $h_i$. To enable early exiting, we add an internal classifier $f_{\theta}$ to each layer of the PLM to make earlier prediction. Denote $X_i=f_\theta(h_i)$ as the output distribution of internal classifier $i$. During training, all of the internal classifiers are trained with the \textit{relevancy loss} $\mathcal{L}_{\text{rel}}$ that maximizes the likelihood of the ground truth $Y$, and the \textit{diversity loss} $\mathcal{L}_{\text{div}}$ that encourages the difference among the internal classifiers. During inference, we sequentially collect the predictions of the passed internal classifiers and employ a combination function $\phi$ to infer the correct label and decide whether or not to exit early. An overview of our approach is shown in Figure~\ref{fig:overview}.

\subsection{Training Objective}
During training, instead of training each internal classifier independently, we attempt to train them as an ensemble. Denote the output distribution of the internal classifier at layer $l$ as $X_l$. At layer $l$, given the joint probability of the first $l$ internal classifiers $X_{1:l}$, our objective is to minimize the probability of error prediction, i.e., 
\begin{equation}
    \min p(\phi(X_{1:l})\neq Y),
\end{equation}

\noindent where $\phi$ is a combination function. The objective to be minimized can be bounded by two inequalities~\cite{fano1961transmission,Hellman1970Probability},
\begin{equation}
    \frac{H(Y)-I(X_{1:l};Y)-1}{\log |Y|} \leq p(\phi(X_{1:l})\neq Y) \leq \frac{H(Y)-I(X_{1:l};Y)}{2},
    \label{eq:ineq}
\end{equation}
where $H(Y)$ is the entropy of variable $Y$, which is defined as $-\sum _yp(y)\log p(y)$. $I$ is mutual information, which measures the amount of information shared by two random variables. The mutual information between $X$ and $Y$ is defined as follows,
\begin{equation}
    I(X;Y) = \sum _X \sum _Y p(xy)\log \frac{p(xy)}{p(x)p(y)}.
\end{equation}
Thus, to minimize the probability of error prediction, the mutual information $I(X_{1:l};Y)$ should be maximized. \cite{Zhou2010Multi} suggest that the mutual information can be decomposed into \textit{relevancy} and \textit{diversity}, which is~\footnote{The diversity defined here is slightly different from \cite{Zhou2010Multi}. See the original paper for detailed derivations.}
\begin{align}
    I(X_{1:l}; Y) &=\sum_{i=1}^lI(X_i; Y) + \mathcal{I}(X_{1:l}|Y) - \mathcal{I}(X_{1:l})\\
    &\geq  \begin{matrix} \underbrace{\sum_{i=1}^lI(X_i; Y)} \\ \mathrm{relevancy} \end{matrix} ~~ \begin{matrix} \underbrace{-\sum_{i=2}^l I(X_i;X_{1:i-1})} \\ \mathrm{diversity} \end{matrix}~~,
\label{eq:mut}
\end{align}
where $\mathcal{I}$ is multi-information, which is the multivariate generalization of mutual information that expresses the dependence among multiple variables. Multi-information is defined as
\begin{equation}
    \mathcal{I}(X_{1:l}) = \sum _{x_{1:l}}p(x_1,\cdots,x_l)\log \frac{p(x_1,\cdots,x_l)}{p(x_1)p(x_2)\cdots p(x_l)}.
\end{equation}

However, directly maximizing mutual information is intractable when the function is parameterized by modern neural networks~\cite{Paninski2003Estimation}. A lower bound that can be optimized in practice is InfoNCE~\cite{Oord2018Repr}, that is
\begin{align}
     I(X_i;Y) &\geq - \mathcal{L}_{\text{InfoNCE}} + \log |\tilde{Y}| \\
     &= \mathop{\mathbb{E}}\limits_{X_i} \left [\log \dfrac{\exp g_\theta (x_i,y_c)}{\sum _{y_{c^\prime }\in \tilde{Y}}\exp g_\theta (x_i,y_{c^\prime})} \right ]+ \log |\tilde{Y}|,
\end{align}
where $\tilde{Y}$ contains a positive label $y_c$ and $|\tilde{Y}|-1$ negative labels drawn from a proposal distribution $p(y)$. Note that $x_i=f(h_i)$ where $h_i$ is the hidden state of the $i$-th layer, thus we can write $g_\theta (x_i,y_j) = g_\theta (f(h_i), y_j)$ as $\tilde{f}_\theta (h_i, y_j)$. In fact, $\tilde{f}_\theta (h_i, y_j)$ is exactly the internal classifier at the $i$-th layer and can be simply implemented as a linear classifier, i.e., $\tilde{f}_\theta (h_i, y_j) = z_{j}^\top W_ih_i$ where $z_j$ is the embedding of the $j$-th label.

When $\tilde{Y} = Y$ and $y_{c^\prime}$ is uniformly sampled, InfoNCE is equivalent to the cross-entropy, therefore minimizing InfoNCE is analogous to minimizing the standard cross-entropy loss~\cite{Kong2020Mutual}. Thus, we can approximately maximize mutual information by minimizing cross-entropy loss
\begin{equation}
    \mathcal{L}_{\text{CE}}(x_i,y) = -\log \dfrac{\exp \tilde{f}_\theta (h_i,y_c)}{\sum _{y_{c^\prime}\in Y}\exp \tilde{f}_\theta (h_i,y_{c^\prime})}.
\end{equation}
Now we have the tool to optimize the relevancy term in Eq.~(\ref{eq:mut}). It is worth noticing that the loss function for the relevancy term is exactly the loss function used in previous work~\cite{Elbayad2020Depth,Schwartz2020Right,Xin2020DeeBERT,Zhou2020BERT}, which we call the \textit{Relevancy Loss}:
\begin{equation}
    \mathcal{L}_{\text{rel}} = \sum _{i=1}^L\mathcal{L}_{\text{CE}}(x_i,y),
\label{eq:relevancy}
\end{equation}


As revealed by Eq.~(\ref{eq:mut}), minimizing the relevancy loss actually maximizes the lower bound of the mutual information between every internal classifier and the ground truth, i.e., $\sum_{i=1}^LI(X_i;Y)$. However, it only optimizes the \textit{relevancy} but not \textit{diversity}. 

Considering the diversity term in Eq.~(\ref{eq:mut}), the estimation of the joint probability $X_{1:i-1}$ is infeasible due to the exponential number of possible values. So we take an approximation following \cite{Zhou2010Multi},\footnote{Another alternative approximation is in \cite{Brown2009Information}.}
\begin{equation}
    I(X_i;X_{1:i-1}) \approx \max _{j<i}I(X_i; X_j).
\end{equation}
To minimize $\max _{j<i}I(X_i; X_j)$ we should maximize $\min _{j<i}\mathcal{L}_{\text{CE}}(x_i, x_j)$. Now it is easy to define the \textit{Diversity Loss}:
\begin{equation}
    \mathcal{L}_{\text{div}} = -\sum _{i=2}^L\min _{j<i}\mathcal{L}_{\text{CE}}(x_i, x_j).
\label{eq:div}
\end{equation}
Combining the relevancy loss (Eq.~(\ref{eq:relevancy})) and the diversity loss (Eq.~(\ref{eq:div})), we have
\begin{align}
    \mathcal{L} & = \mathcal{L}_{\text{rel}} + \mathcal{L}_{\text{div}}\\
    &= \sum _{i=1}^L \mathcal{L}_{\text{CE}}(x_i, y) - \lambda \sum _{i=2}^L\min _{j<i}\mathcal{L}_{\text{CE}}(x_i, x_j),
\label{eq:loss}
\end{align}
where $\lambda$ is a hyperparameter to balance the relevancy loss and the diversity loss. For simplicity, we neglect the weights for different internal classifiers.
By minimizing the loss function above, the lower bound of the mutual information between the joint probability of predictions of all the internal classifiers and the ground truth is maximized, therefore the predictor $\phi$ has more chances to infer the correct label, as revealed by inequalities~(\ref{eq:ineq}). The predictor $\phi$, which is related to the inference strategy, will be discussed latter in Section~\ref{sec:inf}.

We find that the derived loss function is rather intuitive. Considering the loss function applied at layer $i$, which is $\mathcal{L}_{\text{CE}}(x_i,y) - \lambda \min _{j<i} \mathcal{L}_{\text{CE}} (x_i, x_j)$. The first term is the standard maximum-likelihood objective, while the second term is to find a previous internal classifier that has minimum cross-entropy with the current internal classifier and maximize it. Note that maximizing the cross-entropy is analogous to maximizing the Kullback-Leibler (KL) divergence, which measures the difference between two probability distributions. Thus, the second term essentially pulls apart the two distributions $x_i$ and $x_j$ that are the least different therefore promotes the diversity. It is also intuitive that promoting the diversity of the internal classifiers would help ensemble-based inference strategies. In ensemble learning, it is crucial to employ diverse individual learners as members of an ensemble due to the complementarity.

\subsubsection*{Theoretical Analysis}
In this section we take a closer look to understand how this loss function works and provide some insights on setting $\lambda$.

Considering the loss applied at layer $i$, which is $\mathcal{L}_i = \mathcal{L}_{\text{CE}}(x_i,y) - \lambda \min _{j<i} \mathcal{L}_{\text{CE}} (x_i, x_j)$. For simplicity, let $k=\arg \min _{j<i} \mathcal{L}_{\text{CE}} (x_i, x_j)$, $p=x_k$ and $q=x_i$. 

For binary classification, we have
\begin{align}
    \mathcal{L}_i &= \mathcal{L}_{\text{CE}}(q,y) - \lambda \mathcal{L}_{\text{CE}} (q,p) \\
    &= \lambda \left [p_c \log q_c + (1-p_c)\log (1-q_c)\right ] -\log q_c \\
    &= - \left [(1-\lambda p_c)\log q_c + \lambda p_c \log (1-q_c) \right] + \lambda \log (1-q_c) ] \\
    & = \mathcal{L}_{\text{CE}}(q,p^\prime) + \lambda \log (1-q_c),
    \label{eq:bi_anal}
\end{align}
where 
\begin{equation}
    p^\prime (x) =
    \begin{cases}
        1-\lambda p_c&\text{when } x=c\\
        \lambda p_c&\text{when } x\neq c
    \end{cases}
    \label{eq:alpha}
\end{equation}
$c$ is the correct class. The first term in Eq. (\ref{eq:bi_anal}) is analogous to \textit{knowledge distillation} where the teacher distribution is the transformation of distribution of the previous internal classifier $p$. The effect of the first term can also be analogous to \textit{label smoothing}, where the soft target is $\{1-\alpha, \alpha\}$ and $\alpha$ is a pre-defined hyperparameter. In our framework, $\alpha=\lambda p_c$ is dynamic during training. The second term in Eq. (\ref{eq:bi_anal}) is simply enlarging the prediction probability on the correct class.


For multi-class classification, we have
\begin{align}
    \mathcal{L}_i &= \mathcal{L}_{\text{CE}}(q,y) - \lambda \mathcal{L}_{\text{CE}} (q,p) \\
    &= -\log q_c + \lambda \sum _i p_i \log q_i \\
    &= (\lambda p_c - 1)\log q_c + \lambda \sum _{i \neq c} p_i \log q_i
\end{align}
When $\lambda \in (0,1)$, we have $(\lambda p_c -1) < 0$, thus the first term encourages the prediction probability on the correct class $q_c$ to become bigger. The second term promotes the difference in the distribution of $p$ and $q$ at \textit{error} categories. Given the analysis above, we suggest $\lambda \in (0,1)$.

\subsection{Inference (Exiting) Strategy}
\label{sec:inf}
As we optimize our loss function (Eq.~(\ref{eq:loss})), the bound of the prediction error $p(\phi(X_{1:l})\neq Y)$ is minimized. However, whether or not the bound can be reached depends on the ability of the prediction function $\phi$, which is related to our inference strategy to be discussed in this section.

During inference, our objective is to find a function $\phi$ that, at the $l$-th layer, can infer the correct label for a given sample based on the output distributions of the first $l$ internal classifiers, i.e. $X_1, \cdots, X_l$. Rather than using only the last one or more internal classifier in prior work~\cite{Xin2020DeeBERT,Schwartz2020Right,Liu2020FastBERT,Zhou2020BERT}, in this section we attempt to fully exploit information in all of the available internal classifiers.

A straightforward idea is to let all of the past internal classifiers take a vote to get the final prediction. When the number of votes for a certain class reaches a pre-defined threshold, we could believe the model is confident of its prediction and let the sample exit. At layer $l$, we simply count the maximum votes over all of the classes,
\begin{equation}
    V_l = \frac{\max _c\Big\{\sum _{j=1}^l \mathbb{I}(\text{Pred}(x_j)=y_c)\Big\}}{l^{k}},
\label{eq:scale_voting}
\end{equation}
where $1\leq c \leq C$ is the class index, $k\in [0,1)$ is a hyperparameter to balance the importance of each layer. A figure that shows the effect of $k$ on the distribution of $V_l$ can be found in Appendix. The sample is supposed to exit at the $l$-th layer when $V_l \geq \delta$. 



\section{Experiments}
\label{sec:exp}
\begin{table}[t!]
\centering
\resizebox{.8\linewidth}{!}{
\begin{tabular}{l|c|ccccc|c}
\toprule
\multirow{2}{*}{\textbf{Method}}                                               & \textbf{Speed} & \textbf{CoLA}   & \textbf{MRPC}   & \textbf{QQP}    & \textbf{RTE}    & \textbf{SST-2} & \textbf{Macro} \\
                                                                      & \textbf{-up}   & (8.5k) & (3.7K) & (364k) & (2.5K) & (67K) & \textbf{Avg.}  \\ \midrule
\multicolumn{8}{c}{\textit{Dev Set}} \\ \midrule
ALBERT-base~\cite{Lan2020ALBERT}                & 1.0$\times$  & 58.9   & 89.5   & 89.6   & 78.6   & 92.8  & 81.9  \\
ALBERT-6L                                                             & 2.0$\times$  & 53.4   & 85.8   & 86.8   & 73.6   & 89.8  & 77.9  \\
ALBERT-9L                                                             & 1.3$\times$  & 55.2   & 87.1   & 88.3   & 75.9   & 91.3  & 79.6  \\
\midrule
LayerDrop~\cite{Fan2020LayerDrop}               & 2.0$\times$  & 53.6   & 85.9   & 87.3   & 74.3   & 90.7  & 78.4  \\
HeadPrune~\cite{Michel2019Heads}                & 1.2$\times$  & 54.1   & 86.2   & 88.0   & 75.1   & 90.5  & 78.8  \\
\midrule
DeeALBERT~$\dag$~\cite{Xin2020DeeBERT}     & 1.5$\times$  & 57.6   & 89.8   & 89.1   & 79.1   & 92.9  & 81.7  \\
FastALBERT~$\dag$~\cite{Liu2020FastBERT}   & 1.5$\times$  & 58.0   & 89.8   & 89.3   & 79.5   & 92.9  & 81.9  \\
PABEE~\cite{Zhou2020BERT}                       & 1.5$\times$  & 61.2   & 90.0   & 89.6   & 80.1   & 93.0  & 82.8  \\ \midrule
\textit{Ours}                                        &       &        &        &        &        &       &       \\
\ \   w/ Patience                            & 1.5$\times$  & 61.4   & 92.4   & 89.6   & \textbf{80.9}   & 93.2  & 83.5  \\
\ \   w/ Voting                              & 1.5$\times$  & \textbf{61.6}   & \textbf{92.7}   & \textbf{89.8}   & \textbf{80.9}   & \textbf{93.5}  & \textbf{83.7}  \\ \midrule
\multicolumn{8}{c}{\textit{Test Set}}  \\ \midrule
ALBERT-base~$\dag$~\cite{Lan2020ALBERT}    & 1.0$\times$  & 54.1   & 86.9   & 71.1   & 76.4   & 94.0  & 76.5  \\
PABEE~\cite{Zhou2020BERT}                       & 1.5$\times$  & 55.7   & 87.4   & 71.2   & 77.3   & 94.1  & 77.1  \\ \midrule
\textit{Ours}                                        &       &        &        &        &        &       &       \\
\ \   w/ Patience                            & 1.5$\times$  &  \textbf{56.2}      & 87.7   & 71.4       & 77.9   & 94.1  &  77.5     \\
\ \   w/ Voting                              & 1.5$\times$  &  \textbf{56.2}      & \textbf{88.0}   & \textbf{71.5}       & \textbf{78.2}   & \textbf{94.4}  &  \textbf{77.7}    \\
\bottomrule
\end{tabular}
}
\caption{Experimental results with ALBERT backbone on binary text classification tasks. The speed-up ratio is averaged across 5 tasks. $\dag$ denotes the results that are obtained with our implementation, other baseline results are taken from \cite{Zhou2020BERT}.}
\vskip -0.2in
\label{tab:binary}
\end{table}

In this section, we evaluate our proposed training objective and inference strategy on various NLP tasks. Since our objective aims at optimizing multiple injected internal classifiers as an ensemble, it is compatible with any ensemble-based inference strategies, e.g. patience-based strategy\footnote{Patience-based strategy allows a sample to exit when the prediction is unchanged for successive $s$ layers, where $s$ is a hyperparameter called patience to adjust the speed-up during inference.}~\cite{Zhou2020BERT}. Thus, we evaluate our proposed training objective, especially the diversity loss, with both the patience-based and the voting-based inference strategy.

\subsection{Experimental Setup}
We mainly evaluate our approach with ALBERT~\cite{Lan2020ALBERT} due to its superior performance with early exiting (See Figure~\ref{fig:oracle}). Also, ALBERT is the primary backbone used in \cite{Zhou2020BERT}. Nevertheless, to demonstrate the versatility of our training objective with other PLMs, we also conduct experiments with BERT~\cite{Devlin2019BERT} on SST-2 and SST-5.

\paragraph{Tasks and Datasets}
Our experiments are conducted on five binary and three multi-class text classification tasks. For binary classification, we adopt two sentence-pair similarity tasks, MRPC~\cite{Dolan2005MRPC} and QQP~\cite{Wang2019GLUE}, a sentiment classification task, SST-2~\cite{Socher2013Recursive}, a linguistic acceptability task, CoLA~\cite{Warstadt2019CoLA}, and an entailment task, RTE~\cite{Wang2019GLUE}. For multi-class tasks, we adopt a topic classification task, AG's News~\cite{Zhang2015Character}, a sentiment classification task, SST-5~\cite{Socher2013Recursive}, and a question classification task, TREC~\cite{Voorhees1999TREC}. 

\paragraph{Baselines}
We choose the following competitive models as our baselines: (1) The original ALBERT-base. (2) Directly fine-tuning different layers of ALBERT. We fine-tuned 6 and 9 layers of ALBERT with a linear classifier on the top, denoted as ALBERT-6L and ALBERT-9L. (3) Model compression approaches. We compare with LayerDrop~\cite{Fan2020LayerDrop} and HeadPrune~\cite{Michel2019Heads} on ALBERT. (4) Early exiting approaches. We choose DeeBERT~\cite{Xin2020DeeBERT}, FastBERT~\cite{Liu2020FastBERT}, and PABEE~\cite{Zhou2020BERT} as our baselines.

\paragraph{Training} 
The internal classifier is implemented as a MLP with one hidden layer. For each task, we conduct the same grid search for both baselines and ours, then select the one that performs the best on development set. For models trained with diversity loss, we tried $\lambda$ in \{0.1, 0.2, 0.3, 0.5\}. The effect of $\lambda$ is explored in Section~\ref{sec:effect}. Our implementation is based on Hugging Face's Transformers~\cite{Wolf2020Transformers}. All of the experiments are conducted on a single NVIDIA GTX1080Ti GPU. More details about our implementation and best-performed hyperparameters can be found in Appendix.


\paragraph{Inference}
Following prior work~\cite{Teerapittayanon2016BranchyNet,Kaya2019Shallow,Zhou2020BERT}, we test speed of inference in a per-instance setting. The speed-up ratio is roughly defined as the number of original layers divided by the number of layers actually executed. Performances under different speed-up ratios are obtained by adjusting the voting threshold $\delta$. We select $k$ that achieves the best trade-off from \{0, 0.25, 0.5, 0.75\}. The effect of $k$ is explored in Section~\ref{sec:effect}.

\subsection{Main Results}
Experimental results on binary and multi-class text classification tasks are shown in Table~\ref{tab:binary} and Table~\ref{tab:multi-class}, respectively. We re-implement DeeBERT and FastBERT with backbone of ALBERT, denoted as DeeALBERT and FastALBERT. For binary classification tasks, following ~\cite{Zhou2020BERT}, we search over a set of thresholds to find the one achieving the best accuracy while targeting a speed-up ratio between 1.3$\times$ (ALBERT-9L) and 2.0$\times$ (ALBERT-6L). For multi-class tasks, we search the threshold with which the model performs the best while targeting a speed-up ratio between 1.3$\times$ and 2.3$\times$ to better demonstrate the superiority of our approach. The reported speed-up ratio in the tables are averaged over the tasks.  For datasets with more than one metric, we report the mean of the metrics.

\begin{figure}[t]
    \centering
    \begin{subfigure}{0.33\linewidth}
    \centering
    \includegraphics[scale=0.46]{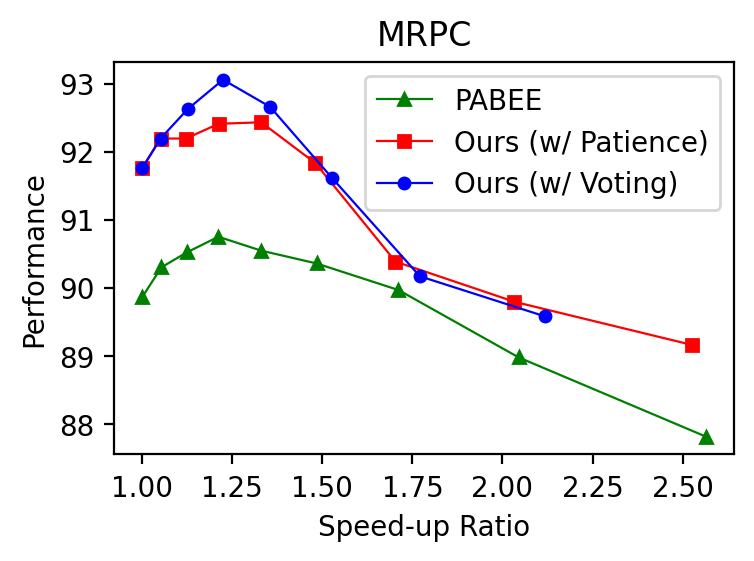}
    \end{subfigure}\hfill
    \begin{subfigure}{0.33\linewidth}
    \centering
    \includegraphics[scale=0.46]{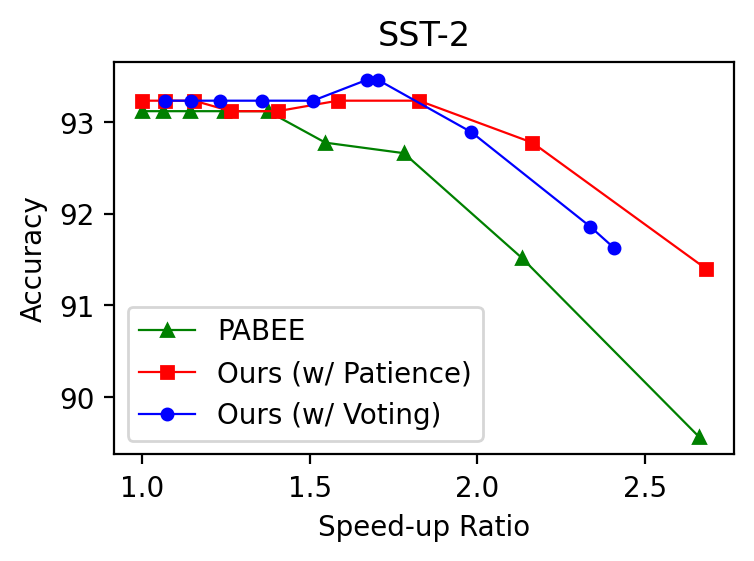}
    \end{subfigure}\hfill
    \begin{subfigure}{0.33\linewidth}
    \centering
    \includegraphics[scale=0.46]{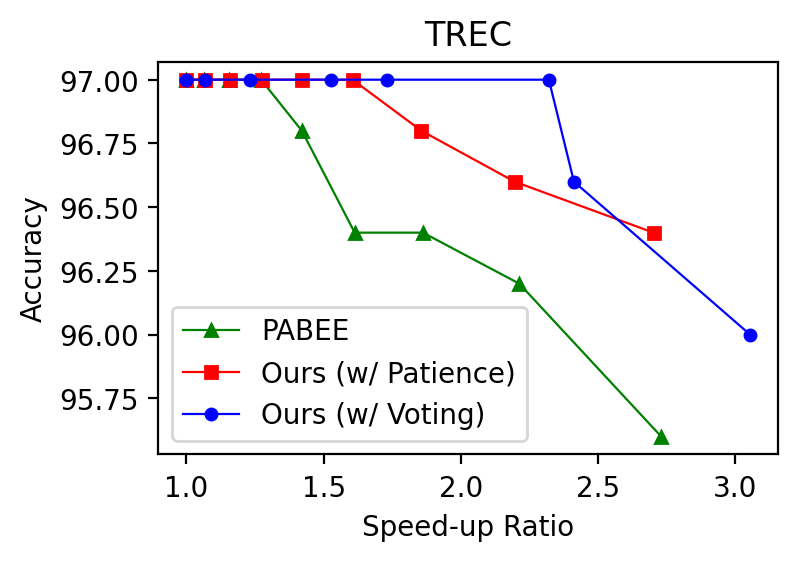}
    \end{subfigure}
    \caption{Performance-speed curves with ALBERT. "Ours" means the model is trained with our loss.}
    \vskip -0.1in
    \label{fig:albert-tradeoff}
\end{figure}

\begin{wrapfigure}{r}{0.4\textwidth}
  \centering
  \vskip -0.2in
  \includegraphics[width=0.4\textwidth]{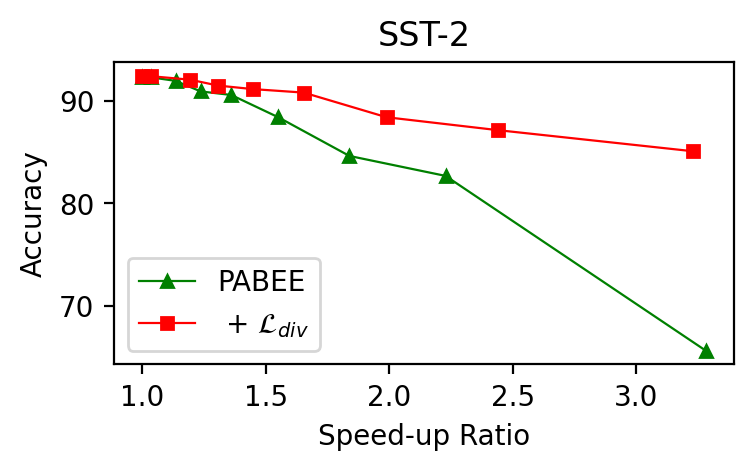}\\
  \includegraphics[width=0.4\textwidth]{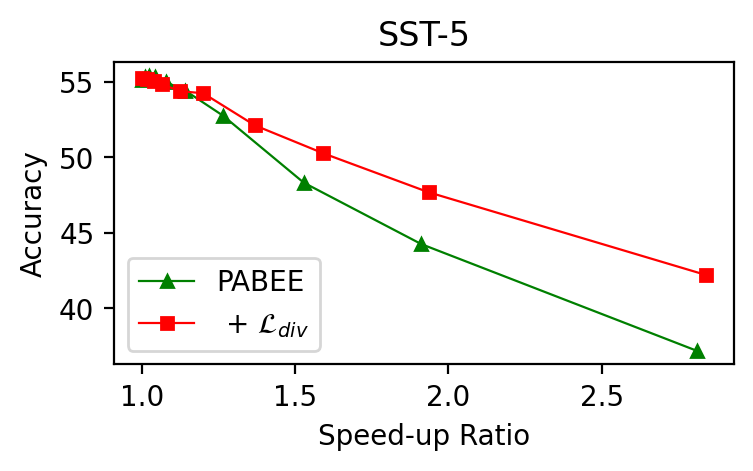}
  \caption{Performance-speed curves with BERT.}
  \vskip -0.5in
  \label{fig:bert-tradeoff}
\end{wrapfigure}

As shown in Figure~\ref{fig:albert-tradeoff}, models trained with our loss function consistently outperform previous state-of-the-art~\cite{Zhou2020BERT} under the same exiting strategy, i.e. patience-based strategy. Equipped with the voting-based strategy, the trade-off can be further improved in most cases. We also confirm the effectiveness of our training objective with backbone of BERT, as shown in Figure~\ref{fig:bert-tradeoff}.

\begin{table}[t]
    \centering
    \resizebox{.7\linewidth}{!}{
    \begin{tabular}{l|c|ccc|c}
    \toprule
    \multirow{2}{*}{\textbf{Method}} & \textbf{Speed} & \textbf{AG's News} & \textbf{SST-5}  & \textbf{TREC} & \textbf{Macro}\\
    & \textbf{-up} & (120K) & (8.5K)  & (5.5K) & \textbf{Avg.}\\
    \midrule
    \# Classes & - & 4 & 5  & 6 & -\\
    \midrule
    ALBERT-base~\cite{Lan2020ALBERT} & 1.0$\times$ & 94.8 & 53.9 & 97.0 & 81.9 \\
    ALBERT-6L   & 2.0$\times$ & 94.5 & 52.6 & 95.8 & 81.0 \\
    ALBERT-9L   & 1.3$\times$ & 94.7 & 53.6 & 97.0 & 81.8 \\
    \midrule
    DeeALBERT~\cite{Xin2020DeeBERT}     & 2.0$\times$ & 94.8 & 53.7 & 96.5 & 81.7\\
    FastALBERT~\cite{Liu2020FastBERT} & 2.0$\times$ & 94.8 & 53.9 & \textbf{97.0} & 81.9\\
    PABEE~\cite{Zhou2020BERT}       & 2.0$\times$ & 94.7 & 54.0 & 96.2 & 81.6 \\
    \midrule
    \textit{Ours} & & & & & \\
    \ w/ Patience & 2.0$\times$ & 95.0 & 54.5 & 96.6 & 82.0 \\
    \ w/ Voting & 2.0$\times$ & \textbf{95.2} & \textbf{54.9} & \textbf{97.0} & \textbf{82.4} \\
    \bottomrule
    \end{tabular}
    }
    \caption{Test results with ALBERT backbone on multi-class text classification tasks. The speed-up ratio is averaged across 3 tasks. The baseline results are obtained with our implementation.}
    \vskip -0.1in
    \label{tab:multi-class}
\end{table}


\subsection{Effect of $\lambda$ and $k$}
\label{sec:effect}
\begin{figure}[t]
    \centering
    \begin{subfigure}{0.49\linewidth}
    \centering
    \includegraphics[scale=0.5]{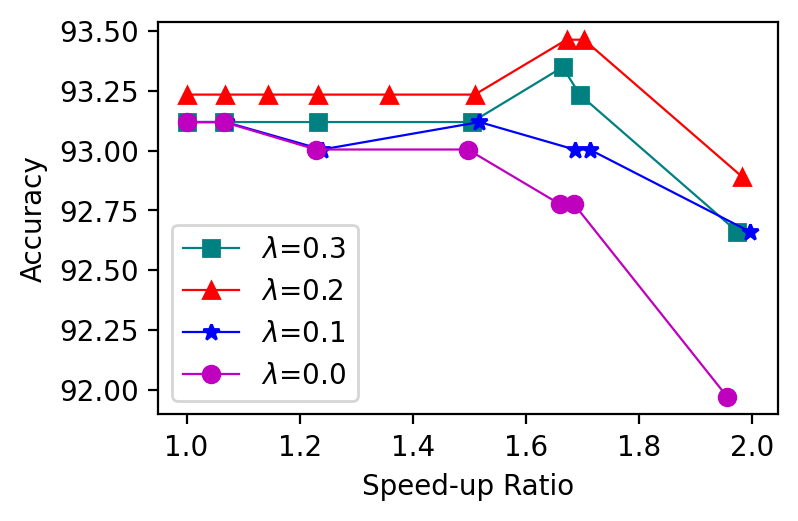}
    \caption{Effect of $\lambda$.}
    \label{fig:lambda}
    \end{subfigure}\hfill
    \begin{subfigure}{0.49\linewidth}
    \centering
    \includegraphics[scale=0.5]{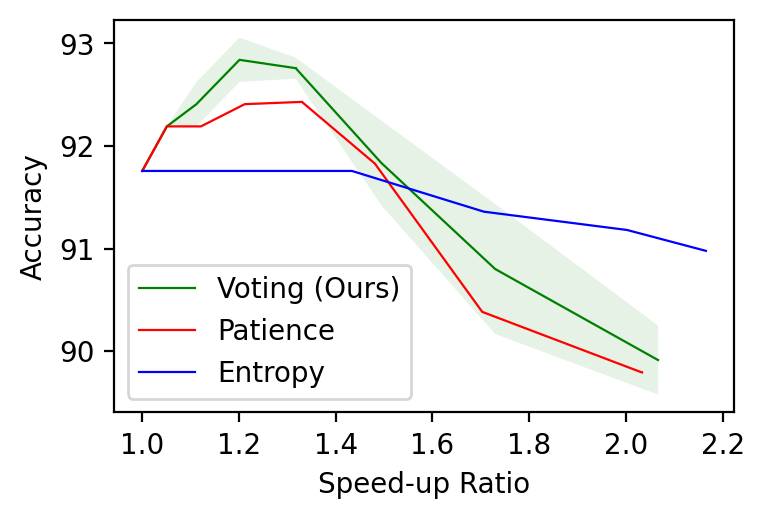}
    \caption{Effect of $k$.}
    \label{fig:k}
    \end{subfigure}
    \caption{Robustness analysis for $\lambda$ and $k$.}
    \vskip -0.1in
    \label{fig:effect}
\end{figure}

In this section we explore the effect of the newly introduced hyperparameters, $\lambda$ in Eq. (\ref{eq:loss}) and $k$ in Eq. (\ref{eq:scale_voting}). First we experiment ALBERT trained using our proposed loss function with $\lambda$ in \{0.1, 0.2, 0.3\}. The experiments are conducted on SST-2 with $\sim$67k training samples. As shown in Figure~\ref{fig:lambda}, the model achieves the best accuracy-speed trade-off in the case of $\lambda =0.2$ while the model without the diversity term (i.e., $\lambda =0$) performs the worst. 

Further, we evaluate model performance and speed-up ratio with different values of $k$ during inference. As shown in Figure~\ref{fig:k}, our voting-based strategy with different $k$ consistently outperforms the patience-based strategy and entropy-based strategy when speed-up ratio is below 1.4$\times$. In addition, we also find that the entropy-based strategy performs surprisingly well when speed-up ratio is above $\sim$1.6$\times$, in which case the accuracy of ensemble-based strategies dramatically drop. Thus, a hybrid exiting strategy is suggested in real-world scenarios.

\subsection{Analysis}
\label{sec:anal}
\begin{figure}[h]
  \centering
  \begin{subfigure}{0.48\linewidth}
    \centering
    \includegraphics[scale=0.5]{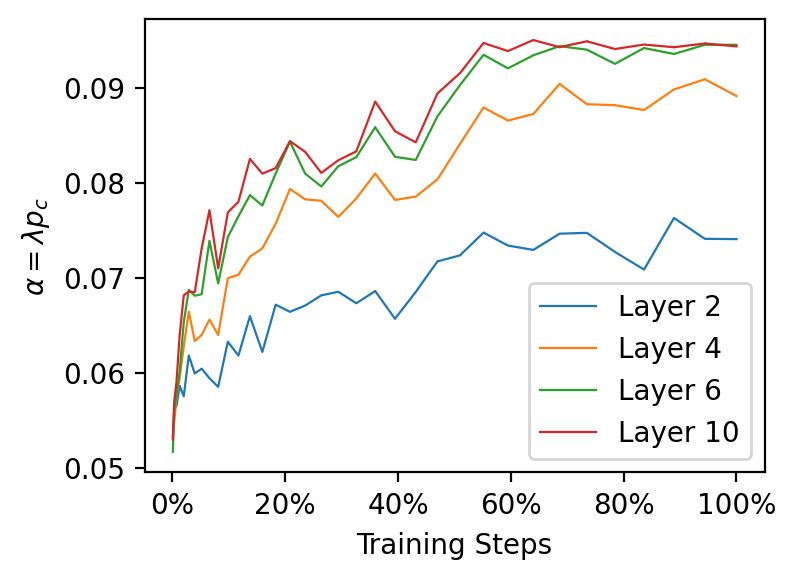}
    \caption{Change of $\alpha=\lambda p_c$ for different layers during training.}
    \label{fig:alpha}
    \end{subfigure}\hfill
    \begin{subfigure}{0.48\linewidth}
    \centering
    \includegraphics[scale=0.53]{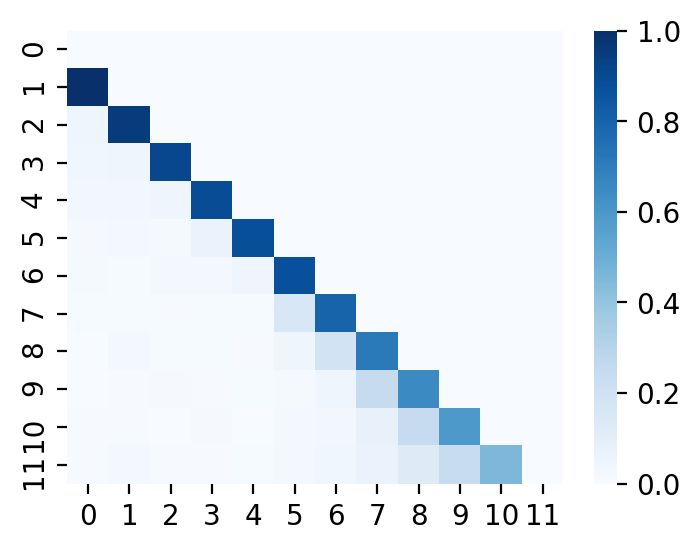}
    \caption{Visualization of percentage of each previous layer that has the closest prediction with each layer.}
    \label{fig:heatmap}
    \end{subfigure}
    \caption{Two empirical analyses on MRPC training set.}
    \vskip -0.1in
\end{figure}

In this section, we attempt to shed some light on the effect of the diversity loss. First we take a closer look at the training procedure with the diversity loss. As revealed in Eq. (\ref{eq:bi_anal}) and Eq. (\ref{eq:alpha}), the diversity loss has a similar effect as label smoothing in the case of binary classification. In label smoothing, the hard target $\{1,0\}$ is softened into a static soft target $\{1-\alpha, \alpha\}$. In our training objective, the distribution of each internal classifier is trained with a soft target $\{1-\lambda p_c, \lambda p_c\}$. Thus, $\lambda p_c$ can be analogous to a dynamic $\alpha$ in label smoothing. We draw the change of $\alpha=\lambda p_c$ for different layers during training on MRPC in Figure~\ref{fig:alpha}. We can find that the dynamic targets for deep layers are softer than shallow layers. In addition, targets for deeper layers converge to $\lambda=0.1$ faster. We conjecture that this helps alleviate the overconfident problem in deep layers~\cite{Guo2017Calibration}. Second we investigate which previous layer has the closest prediction to the current layer. In Figure~\ref{fig:heatmap} we show the percentage of each previous layer that has the closest prediction to each layer over all the training steps. It is natural to find that adjacent layers are highly correlated therefore have the most similar prediction. Thus, the diversity loss mainly acts on adjacent layers, which provides an approximation to reduce the computation of searching for the previous layer that has minimum cross entropy with the current layer.

\section{Discussion}
\label{sec:discussion}
This paper is motivated by the tempting capability of early exiting that can achieve faster and more accurate inference. To take a step toward the oracle model, we provide a general formulation for early exiting from the perspective of ensemble learning and information theory. Under our formulation, we propose a novel loss function and a voting-based inference strategy. The effectiveness of the both contributions are confirmed through experiments.

Our proposed methods attempt to achieve efficient inference, which can reduce the computational cost and carbon emission, and can facilitate the deployment of huge models on mobile devices and real-time scenarios. As a plug-in technique, our proposed approach would not introduce new potential negative societal impacts. Though, more work is needed to investigate the effect of early exiting models on social bias such as gender bias.

\section{Limitations}
\label{sec:limit}
There are two main limitations in this work. \textbf{First}, we introduce two new hyperparameters, i.e. $\lambda$ and $k$, which increase the computation cost. During training, one should add $\lambda$ to the hyperparameter search, which may cost more search time. During inference, one should try different values of $k$ to get a satisfying trade-off. Nevertheless, the increased search cost may not be too much according to the robustness analysis in Section~\ref{sec:effect}. \textbf{Second}, our approach is mainly evaluated with ALBERT, therefore whether it is applicable to a wider range of backbones is still unknown, though in Figure~\ref{fig:bert-tradeoff} we show that the proposed diversity loss also works for BERT.



\bibliography{ref}
\bibliographystyle{unsrt}

\newpage
\appendix

\section{Appendix}

\subsection{Voting Values}
\begin{figure}[h]
  \centering
  \begin{subfigure}{0.24\textwidth}
    \centering
    \begin{tikzpicture}[scale=0.4]
      \begin{axis}[
        xlabel=Layer,
        ylabel=$V_l$,
        grid=major,
      ]
      \addplot[scatter, only marks]  coordinates{
        (1, 1.0)
        (2, 1.0)
        (2, 2.0)
        (3, 2.0)
        (3, 3.0)
        (4, 2.0)
        (4, 3.0)
        (4, 4.0)
        (5, 3.0)
        (5, 4.0)
        (5, 5.0)
        (6, 3.0)
        (6, 4.0)
        (6, 5.0)
        (6, 6.0)
        (7, 4.0)
        (7, 5.0)
        (7, 6.0)
        (7, 7.0)
        (8, 4.0)
        (8, 5.0)
        (8, 6.0)
        (8, 7.0)
        (8, 8.0)
        (9, 5.0)
        (9, 6.0)
        (9, 7.0)
        (9, 8.0)
        (9, 9.0)
        (10, 5.0)
        (10, 6.0)
        (10, 7.0)
        (10, 8.0)
        (10, 9.0)
        (10, 10.0)
        (11, 6.0)
        (11, 7.0)
        (11, 8.0)
        (11, 9.0)
        (11, 10.0)
        (11, 11.0)
        (12, 6.0)
        (12, 7.0)
        (12, 8.0)
        (12, 9.0)
        (12, 10.0)
        (12, 11.0)
        (12, 12.0)
      };
      \end{axis}
      \end{tikzpicture}
      \caption{$k=0$}
  \end{subfigure}
  \hfill
  \begin{subfigure}{0.24\textwidth}
    \centering
    \begin{tikzpicture}[scale=0.4]
      \begin{axis}[
        xlabel=Layer,
        ylabel=$V_l$,
        grid=major,
      ]
      \addplot[scatter, only marks]  coordinates{
        (1, 1.0)
        (2, 0.8408964152537146)
        (2, 1.6817928305074292)
        (3, 1.519671371303185)
        (3, 2.279507056954778)
        (4, 1.414213562373095)
        (4, 2.1213203435596424)
        (4, 2.82842712474619)
        (5, 2.006220914929266)
        (5, 2.674961219905688)
        (5, 3.34370152488211)
        (6, 1.9168293127388174)
        (6, 2.5557724169850897)
        (6, 3.1947155212313625)
        (6, 3.833658625477635)
        (7, 2.4591526118050577)
        (7, 3.073940764756322)
        (7, 3.6887289177075866)
        (7, 4.303517070658851)
        (8, 2.378414230005442)
        (8, 2.973017787506803)
        (8, 3.5676213450081633)
        (8, 4.162224902509524)
        (8, 4.756828460010884)
        (9, 2.886751345948129)
        (9, 3.464101615137755)
        (9, 4.041451884327381)
        (9, 4.618802153517007)
        (9, 5.196152422706632)
        (10, 2.8117066259517456)
        (10, 3.3740479511420944)
        (10, 3.9363892763324437)
        (10, 4.498730601522793)
        (10, 5.061071926713142)
        (10, 5.623413251903491)
        (11, 3.2946029206566747)
        (11, 3.843703407432787)
        (11, 4.3928038942089)
        (11, 4.9419043809850125)
        (11, 5.4910048677611245)
        (11, 6.040105354537237)
        (12, 3.2237097954706257)
        (12, 3.760994761382397)
        (12, 4.298279727294168)
        (12, 4.835564693205939)
        (12, 5.37284965911771)
        (12, 5.910134625029481)
        (12, 6.4474195909412515)
      };
      \end{axis}
      \end{tikzpicture}
      \caption{$k=0.25$}
  \end{subfigure}
  \hfill
  \begin{subfigure}{0.24\textwidth}
    \centering
    \begin{tikzpicture}[scale=0.4]
      \begin{axis}[
        xlabel=Layer,
        ylabel=$V_l$,
        grid=major,
      ]
      \addplot[scatter, only marks]  coordinates{
        (1, 1.0)
        (2, 0.7071067811865475)
        (2, 1.414213562373095)
        (3, 1.1547005383792517)
        (3, 1.7320508075688774)
        (4, 1.0)
        (4, 1.5)
        (4, 2.0)
        (5, 1.3416407864998738)
        (5, 1.7888543819998317)
        (5, 2.23606797749979)
        (6, 1.2247448713915892)
        (6, 1.6329931618554523)
        (6, 2.041241452319315)
        (6, 2.4494897427831783)
        (7, 1.5118578920369088)
        (7, 1.889822365046136)
        (7, 2.2677868380553634)
        (7, 2.6457513110645903)
        (8, 1.414213562373095)
        (8, 1.7677669529663687)
        (8, 2.1213203435596424)
        (8, 2.4748737341529163)
        (8, 2.82842712474619)
        (9, 1.6666666666666667)
        (9, 2.0)
        (9, 2.3333333333333335)
        (9, 2.6666666666666665)
        (9, 3.0)
        (10, 1.5811388300841895)
        (10, 1.8973665961010275)
        (10, 2.2135943621178655)
        (10, 2.5298221281347035)
        (10, 2.846049894151541)
        (10, 3.162277660168379)
        (11, 1.8090680674665818)
        (11, 2.1105794120443453)
        (11, 2.412090756622109)
        (11, 2.7136021011998728)
        (11, 3.015113445777636)
        (11, 3.3166247903554)
        (12, 1.7320508075688774)
        (12, 2.0207259421636903)
        (12, 2.3094010767585034)
        (12, 2.598076211353316)
        (12, 2.886751345948129)
        (12, 3.1754264805429417)
        (12, 3.464101615137755)
      };
      \end{axis}
      \end{tikzpicture}
      \caption{$k=0.5$}
  \end{subfigure}
  \hfill
  \begin{subfigure}{0.24\textwidth}
    \centering
    \begin{tikzpicture}[scale=0.4]
      \begin{axis}[
        xlabel=Layer,
        ylabel=$V_l$,
        grid=major,
      ]
      \addplot[scatter, only marks]  coordinates{
        (1, 1.0)
        (2, 0.5946035575013605)
        (2, 1.189207115002721)
        (3, 0.8773826753016617)
        (3, 1.3160740129524926)
        (4, 0.7071067811865475)
        (4, 1.0606601717798212)
        (4, 1.414213562373095)
        (5, 0.8972092687327323)
        (5, 1.1962790249769764)
        (5, 1.4953487812212205)
        (6, 0.7825422900366437)
        (6, 1.0433897200488582)
        (6, 1.3042371500610728)
        (6, 1.5650845800732873)
        (7, 0.9294723209701633)
        (7, 1.1618404012127042)
        (7, 1.394208481455245)
        (7, 1.6265765616977856)
        (8, 0.8408964152537146)
        (8, 1.0511205190671433)
        (8, 1.2613446228805718)
        (8, 1.4715687266940005)
        (8, 1.6817928305074292)
        (9, 0.9622504486493763)
        (9, 1.1547005383792515)
        (9, 1.3471506281091268)
        (9, 1.539600717839002)
        (9, 1.7320508075688772)
        (10, 0.8891397050194614)
        (10, 1.0669676460233537)
        (10, 1.2447955870272458)
        (10, 1.4226235280311381)
        (10, 1.6004514690350304)
        (10, 1.7782794100389228)
        (11, 0.993360156457021)
        (11, 1.1589201825331912)
        (11, 1.3244802086093614)
        (11, 1.4900402346855317)
        (11, 1.6556002607617017)
        (11, 1.821160286837872)
        (12, 0.9306048591020996)
        (12, 1.0857056689524496)
        (12, 1.2408064788027995)
        (12, 1.3959072886531494)
        (12, 1.5510080985034993)
        (12, 1.7061089083538492)
        (12, 1.8612097182041991)
      };
      \end{axis}
      \end{tikzpicture}
      \caption{$k=0.75$}
  \end{subfigure}
  \caption{Values of $V_l$ when $k$ takes different values. Given a threshold $\delta$, only points with values beyond it can exit. For example, when $\delta=1.5$ when $k=0.75$, the sample could exit at layer 6 if it gets 6 votes and at layer 9 if it gets 8 or 9 votes, and so on.}
  \label{fig:vote}
\end{figure}
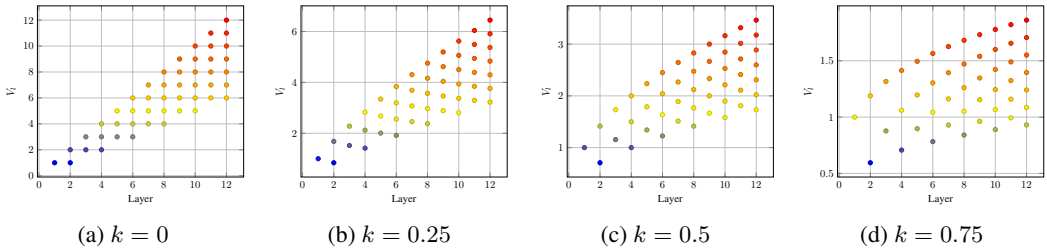

The distribution of values of $V_l$ in Eq. (\ref{eq:scale_voting}) is important for selecting the early exiting threshold $\delta$. We show the distributions of $V_l$ when $k=\{0, 0.25, 0.5, 0.75\}$ in the context of binary classification in Figure~\ref{fig:vote}.

\subsection{Comparison of Exiting Strategies}
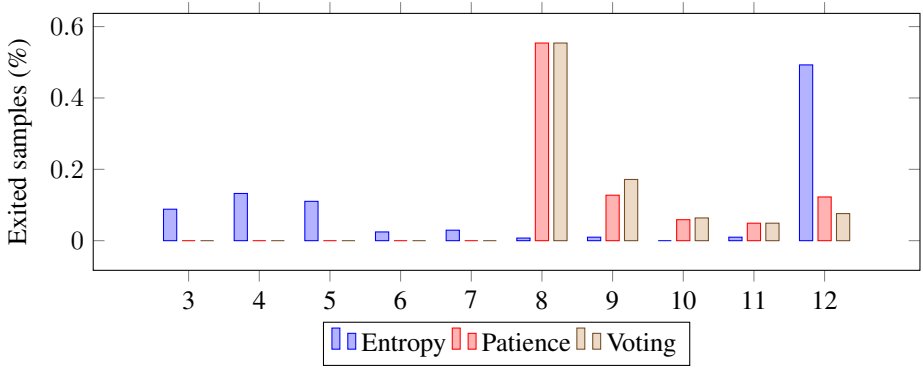
\begin{figure}[h]
    \centering
   \begin{tikzpicture}
\begin{axis}[
    ybar,
    enlargelimits=0.15,
    legend style={at={(0.5,-0.2)},
      anchor=north,legend columns=-1},
    ylabel={Exited samples (\%)},
    symbolic x coords={3,4,5,6,7,8,9,10,11,12},
    xtick=data,
    bar width=5pt,
    height=5cm,
    width=.9\linewidth,
    ]
\addplot coordinates {
(3,0.08823529411764706) 
(4,0.1323529411764706) 
(5,0.11029411764705882) 
(6,0.024509803921568627) 
(7,0.029411764705882353) 
(8,0.007352941176470588) 
(9,0.00980392156862745)
(10, 0)
(11,0.00980392156862745) 
(12,0.49264705882352944) };
\addplot coordinates {
(3,0.0) 
(4,0.0) 
(5,0.0) 
(6,0.0) 
(7,0.0) 
(8,0.553921568627451) 
(9,0.12745098039215685) 
(10,0.058823529411764705) 
(11,0.049019607843137254) 
(12,0.12254901960784313) };
\addplot coordinates {
(3,0.0) 
(4,0.0) 
(5,0.0) 
(6,0.0) 
(7,0.0) 
(8,0.553921568627451) 
(9,0.1715686274509804) 
(10,0.06372549019607843) 
(11,0.049019607843137254) 
(12,0.07598039215686274) };
\legend{Entropy,Patience,Voting}
\end{axis}
\end{tikzpicture}
    \caption{Number of samples that correctly exit at each layer.}
    \label{fig:exit_ratio}
\end{figure}

In Figure~\ref{fig:k} we have shown the trade-off curves achieved by the three exiting strategies, in which we show that the voting-based strategy performs the best at low speed-up while the simple entropy-based strategy performs well when the speed-up ratio becomes higher. To take a closer look, in Figure~\ref{fig:exit_ratio} we shown the distribution of samples that correctly exited at each layer. The results are obtained under the three different exiting strategies with the same speed-up ratio ($\sim$1.30$\times$). We can find that under ensemble-based strategies (patience-based and voting-based) most samples exited at middle layers while more samples exited at earlier layers under entropy-based strategy.

\subsection{Visualization of Hidden States}
\begin{figure}[h]
    \centering
    \includegraphics[width=\textwidth]{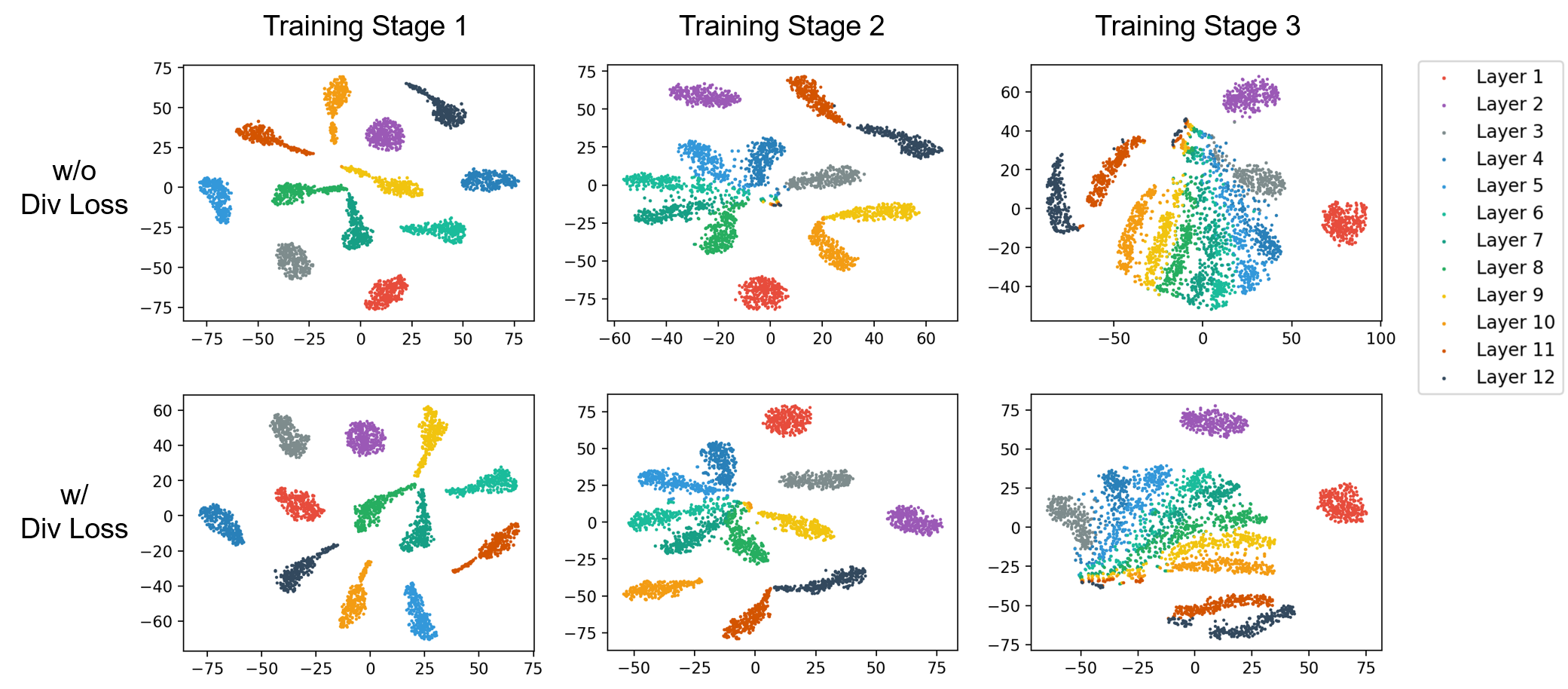}
    \caption{Visualization of hidden states of each layer at different training steps.}
    \label{fig:hidden}
\end{figure}

We also visualize the hidden states of each layer during training with and without our proposed diversity loss. The visualization is shown in Figure~\ref{fig:hidden}, in which we use t-SNE to plot two-dimensional hidden states of each layer during training on SST-2. We only plot smaples with positive labels. The three training stages are 100, 1k, and 10k training steps, respectively. As shown in the Figure, the correlation of the hidden states of adjacent layers increases with training, therefore the representations of adjacent layers become closer. 

\subsection{Implementation Details}
As shown in Eq.~(\ref{eq:loss}), our loss function is a simple sum instead of a weighted sum of the losses of each internal classifier. Since the relevancy loss (Eq.~(\ref{eq:relevancy})) that assigns each internal classifier different weights has proven to perform well in prior work~\cite{Schwartz2020Right,Zhou2020BERT}, we also explored using the weighted sum of our proposed loss for each internal classifier, i.e.,
\begin{equation}
    \mathcal{L} = \sum _{i=1}^L \alpha _i\mathcal{L}_{\text{CE}}(x_i, y) -  \sum _{i=2}^L \beta _i\min _{j<i}\mathcal{L}_{\text{CE}}(x_i, x_j).
\end{equation}

Our preliminary experiments show that the simply summed loss function performs well with ALBERT but poorly with BERT. Thus, we follow \cite{Zhou2020BERT} to set $\alpha_i=i$ for BERT. Further, we find that better results can sometimes be obtained when removing the diversity term of the last internal classifier (i.e., $\beta _L =0$).

\subsection{Dataset Statistics}
The task statistics are listed in Table~\ref{tab:data}.

\begin{table}[h]
\centering
\begin{tabular}{lrrrl}
\toprule
\multicolumn{1}{l}{\textbf{Dataset}} & \multicolumn{1}{r}{\textbf{Classes}} & \multicolumn{1}{r}{\textbf{$|$Train$|$}} & \multicolumn{1}{r}{\textbf{$|$Test$|$}} & \multicolumn{1}{l}{\textbf{Task}} \\
\midrule
MRPC & 2 & 3.7k & 1.7k & Similarity \\
QQP & 2 & 364k & 391k & Similarity \\
CoLA & 2 & 8.5k & 1k & Acceptability \\
RTE & 2 & 2.5k & 3k & Entailment \\
SST-2 & 2 & 67k & 1.8k & Sentiment \\
\midrule
AG's News & 4 & 120k & 7.6k & Topic \\
SST-5 & 5 & 8.5k & 2.2k & Sentiment \\
TREC & 6 & 5.5k & 0.5k & Question \\                 
\bottomrule
\end{tabular}
\caption{Statistics of our experimented datasets.}
\label{tab:data}
\end{table}

\subsection{Hyperparameters}
For downstream tasks, we perform grid search over batch sizes of \{16, 32\}, learning rate of \{1e-5, 2e-5, 3e-5, 5e-5\}, training epochs of \{3, 4, 5\}, and $\lambda$ of \{0.1, 0.2, 0.3, 0.5\} with an Adam optimizer. For each group of hyperparameters, we perform 3 search trials. The maximum sequence length is set to 128 for all tasks. For RTE, to achieve comparable results to prior work, we follow~\cite{Lan2020ALBERT} to start fine-tuning from the checkpoint trained on MNLI. The hyperparameters that perform the best on development set are listed in Table~\ref{tab:hyper}.
\begin{table}[h]
    \centering
    \begin{tabular}{lccccc}
    \toprule
    \textbf{\textbf{Taks}} & \textbf{\textbf{LR}} & \textbf{\textbf{BSZ}} & \textbf{\textbf{Epoch}} & \textbf{\textbf{WSR}} & \textbf{\textbf{$\lambda$}} \\
    \midrule
    MRPC          & 5e-5                   & 32                  & 3              & 0.1             & 0.1             \\
    SST-2         & 1e-5                   & 16                  & 5              & 0.0             & 0.5             \\
    QNLI          & 2e-5                   & 32                  & 3              & 0.1             & 0.5             \\
    RTE           & 3e-5                   & 16                  & 5              & 0.1             & 0.1             \\
    AG's News     & 1e-5                   & 32                  & 5              & 0.1             & 0.5             \\
    TREC          & 2e-5                   & 16                  & 4              & 0.1             & 0.3             \\
    SST-5         & 2e-5                   & 16                  & 3              & 0.2             & 0.1     \\
    \bottomrule
    \end{tabular}
    \caption{Hyperparameters used in our experiments. LR: Learning Rate. BSZ: Batch Size. WSR: Warmup Step Ratio.}
    \label{tab:hyper}
\end{table}



\end{document}